\documentclass[letterpaper]{article} 
\usepackage{aaai2026} 
\usepackage{times}  
\usepackage{helvet}  
\usepackage{courier}  
\usepackage[hyphens]{url}  
\usepackage{graphicx} 
\usepackage{natbib}  
\usepackage{caption} 
\usepackage{algorithm}
\usepackage{algorithmic}
\usepackage{amsmath}  
\usepackage{amssymb} 
\usepackage{multirow}
\usepackage{booktabs}
\usepackage{tabularx}
\usepackage{newfloat}
\usepackage{listings}
\DeclareCaptionStyle{ruled}{labelfont=normalfont,labelsep=colon,strut=off} 
\floatstyle{ruled}
\newfloat{listing}{tb}{lst}{}
\floatname{listing}{Listing}
\usepackage{makecell}

\usepackage{xcolor}
\usepackage{pifont}
\newcommand{\cmark}{\textcolor{green!60!black}{\ding{51}}}  
\newcommand{\xmark}{\textcolor{red}{\ding{55}}}             

\title{AIS-LLM: A Unified Framework for Maritime Trajectory Prediction, Anomaly Detection, and Collision Risk Assessment with Explainable Forecasting}
\author {
    Hyobin Park\textsuperscript{\rm 1}\equalcontrib,
    Jinwook Jung\textsuperscript{\rm 1}\equalcontrib,
    Minseok Seo\textsuperscript{\rm 2,4},
    Hyunsoo Choi\textsuperscript{\rm 3},
    Deukjae Cho\textsuperscript{\rm 3},
    Sekil Park\textsuperscript{\rm 3}\thanks{Corresponding author.},
    Dong-Geol Choi\textsuperscript{\rm 1,2}\footnotemark[1]
}

\affiliations {
    \textsuperscript{\rm 1} Hanbat National University, South Korea \\
    \textsuperscript{\rm 2} Korea Advanced Institute of Science and Technology (KAIST) \\
    \textsuperscript{\rm 3} KRISO, South Korea \\
    \textsuperscript{\rm 4}  ANTLAB, South Korea \\
    \{hyobin.park, jinwook.jung\}@edu.hanbat.ac.kr,
    msseok96@gmail.com,
    \{troychoi, djcho, skpark\}@kriso.re.kr,
    dgchoi@hanbat.ac.kr
}

\begin{document}

\maketitle

\begin{abstract}
With the increase in maritime traffic and the mandatory implementation of the Automatic Identification System (AIS), the importance and diversity of maritime traffic analysis tasks based on AIS data, such as vessel trajectory prediction, anomaly detection, and collision risk assessment, is rapidly growing. However, existing approaches tend to address these tasks individually, making it difficult to holistically consider complex maritime situations. To address this limitation, we propose a novel framework, AIS-LLM, which integrates time-series AIS data with a large language model (LLM). AIS-LLM consists of a Time-Series Encoder for processing AIS sequences, an LLM-based Prompt Encoder, a Cross-Modality Alignment Module for semantic alignment between time-series data and textual prompts, and an LLM-based Multi-Task Decoder. This architecture enables the simultaneous execution of three key tasks: trajectory prediction, anomaly detection, and risk assessment of vessel collisions within a single end-to-end system. Experimental results demonstrate that AIS-LLM outperforms existing methods across individual tasks, validating its effectiveness. Furthermore, by integratively analyzing task outputs to generate situation summaries and briefings, AIS-LLM presents the potential for more intelligent and efficient maritime traffic management.
\end{abstract}


\section{Introduction}

\begin{table*}[ht]
\centering
\small
\begin{tabular}{lccccc}
\toprule
Model & \makecell[c]{Trajectory\\Prediction} & \makecell[c]{Anomaly\\Detection} & \makecell[c]{Collision Risk\\Assessment} & \makecell[c]{Multi-task\\Capability} & \makecell[c]{Explainable\\Capability} \\

\midrule
\citet{li2023ship}               & \cmark & \xmark & \xmark & \xmark & \xmark \\
\citet{korupoju2025ship}         & \xmark & \xmark & \cmark & \xmark & \xmark \\
\citet{tritsarolis2023collision} & \xmark & \xmark & \cmark & \xmark & \xmark \\
\citet{nguyen2018multi}          & \xmark & \cmark & \xmark & \cmark & \xmark \\
\citet{zissis2015cloud}          & \cmark & \cmark & \xmark & \cmark & \xmark \\
\textbf{\textbf{AIS-LLM (Ours)}}         & \cmark & \cmark & \cmark & \cmark & \cmark \\
\bottomrule
\end{tabular}
\caption{Comparison of capabilities between existing maritime traffic analysis models and our proposed AIS-LLM. While existing models can only perform one or two tasks among trajectory prediction, anomaly detection, and collision risk assessment, our proposed AIS-LLM is the only framework that integrates all tasks while providing interpretable natural language explanations.}
\label{tab:comparison_cap}
\end{table*}

With the rapid growth of maritime traffic, the importance of maritime traffic analysis using Automatic Identification System (AIS) data is increasing. AIS devices regularly transmit real-time information such as vessel position, speed, and course, which has enabled the development of various technologies for improving maritime safety and efficiency, including vessel trajectory prediction~\cite{shin2024deep, li2023ship, zhang2022vessel, jiao2025multi}, abnormal behavior detection~\cite{nguyen2018multi, zissis2015cloud, fu2017finding,zhang2020analysis,stach2023maritime}, and collision risk assessment~\cite{tritsarolis2023collision, korupoju2025ship, tengesdal2021ship, marino2023new}. However, existing studies suffer from the following limitations.

\begin{figure}[t] 
    \centering
    \includegraphics[width=0.99\linewidth]{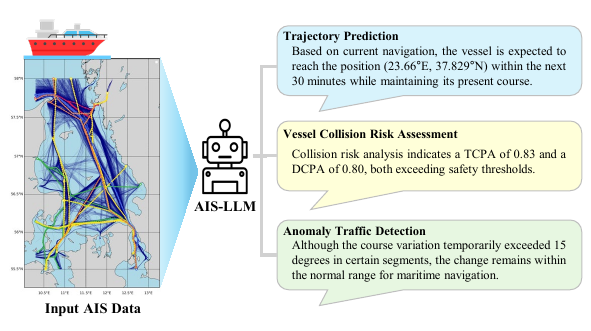}
    \caption{\textbf{Overview of the AIS-LLM Framework.} The system takes maritime AIS data as input (left) and performs three key maritime traffic analysis tasks simultaneously through an integrated end-to-end architecture: (1) \textbf{Trajectory Prediction} – predicts vessel trajectories over a given time horizon; (2) \textbf{Vessel Collision Risk Assessment} – quantifies the likelihood of collisions and estimates a collision risk index; (3) \textbf{Anomaly Traffic Detection} – identifies abnormal vessel behaviors. All prediction results are accompanied by interpretable explanations in natural language.}
    \label{fig:intro_teaser}
\end{figure}

First, there is a limitation in the task-specific analytical approach. Prior works have treated trajectory prediction, anomaly detection, and collision risk assessment as independent problems. Such siloed approaches hinder integrated understanding of maritime traffic situations and fail to effectively support complex decision-making in real-world traffic control scenarios. These tasks, however, are closely related, and an integrated approach is essential.

Second, there is the issue of poor interpretability of prediction results. Deep learning-based models for maritime traffic analysis~\cite{nguyen2021geotracknet, nguyen2021traisformer, wang2025model} show high predictive accuracy, but they output predictions merely as numerical values derived from AIS data consisting of various interrelated variables (e.g., position, time, speed, direction). These outputs are difficult to interpret intuitively and limit the range of actionable insights. In maritime domains, it is essential not only to provide numerical results but also to reflect the situational context of various variables. A lack of interpretability can reduce the acceptability of predictions, posing challenges to real-world adoption. Therefore, explainability is crucial for traffic controllers and vessel operators to understand and trust model outputs.

Third, current models lack sufficient modeling of interactions among multiple vessels. Most existing approaches analyze the behavior of individual vessels in isolation and fail to account for the vessel-to-vessel interactions critical in complex maritime scenarios. Understanding such interactions is vital for safe navigation and efficient traffic flow, but prior models fall short in comprehensively capturing these dynamics.

Table~\ref{tab:comparison_cap} compares existing maritime traffic analysis models with our proposed AIS-LLM framework. Most existing models handle only one or two tasks—trajectory prediction, anomaly detection, or collision risk assessment—while our proposed AIS-LLM performs all three tasks in an integrated manner and also generates explainable outputs in natural language.

To address these limitations, we propose a novel end-to-end framework called \textbf{Automatic Identification System-Large Language Model (AIS-LLM)}, as illustrated in Figure~\ref{fig:intro_teaser}. AIS-LLM combines the strong reasoning and understanding capabilities of large language models (LLMs) with techniques for processing time-series data, enabling integrated vessel trajectory prediction, anomaly detection, and collision risk assessment.

The main contributions of this work are as follows:

\begin{itemize}
    \item \textbf{Unified Maritime Traffic Analysis Framework:} We propose an end-to-end system that simultaneously performs trajectory prediction, anomaly detection, and collision risk assessment using AIS data. Our framework enables comprehensive analysis by modeling not only individual vessels but also interactions among multiple vessels.

    \item \textbf{LLM-Based Explainable Time-Series Forecasting:} We design a structure that can process the diverse variables in AIS time-series data and provide natural language explanations that include not only predicted positions but also contextual information such as speed, direction, collision risk with other vessels, and anomaly status. These explanations achieve competitive performance in zero-shot maritime briefing tasks.

    \item \textbf{Multi-Task Learning Architecture:} We design a multi-task learning framework that shares knowledge across various maritime analysis tasks, maximizing the benefits of transfer learning to improve performance and generalization simultaneously.
\end{itemize}


\section{Related Work}

\subsection{Vessel Trajectory Prediction}
Vessel trajectory prediction is a fundamental task in maritime applications such as route planning, collision avoidance, and port traffic optimization. Early approaches were primarily based on statistical models like Kalman filters and hidden Markov models, which were later surpassed by deep learning methods employing recurrent neural networks (e.g., LSTM, GRU). More recently, models such as TrAISformer~\cite{nguyen2024transformer}, AIS-ACNet~\cite{shin2024deep}, and multi-factor deep learning approaches~\cite{jiao2025multi} have leveraged Transformer architectures, auxiliary objectives, and dynamic AIS features such as speed over ground (SOG) and course over ground (COG) to more effectively capture complex spatiotemporal dependencies in AIS data.

While these methods have achieved notable improvements in predictive accuracy, a critical limitation remains in their lack of interpretability. Most existing models focus on generating future trajectories as sequences of positions without providing explanations for why a particular trajectory was predicted or how it relates to the broader maritime context, such as navigational constraints, traffic patterns, or operational intent. This lack of semantic grounding limits their utility in real-world scenarios, where decision-makers require not only accurate forecasts but also contextual insights to support situational awareness and risk assessment. Therefore, there is a growing need for trajectory prediction frameworks that combine quantitative accuracy with explainable, context-aware reasoning.

\subsection{Vessel Risk Assessment}
Vessel collision risk assessment using AIS data is a critical component in ensuring maritime safety. It typically involves quantifying the likelihood of collision between vessels based on kinematic features such as distance, speed, and bearing, often expressed through a Collision Risk Index (CRI). Traditional approaches leverage features like the Distance to Closest Point of Approach (DCPA) and Time to Closest Point of Approach (TCPA) as input to classical machine learning models such as SVM, RVM, and MLP~\cite{tritsarolis2023collision, park2021estimation, tritsarolis2022vessel}. More recent developments employ deep learning techniques, including Convolutional LSTM architectures, to estimate region-specific collision probabilities using spatiotemporal AIS and weather data~\cite{korupoju2025ship}.

Despite these advancements, most existing approaches handle collision risk prediction as an isolated task, lacking integration with broader maritime situational contexts. They often operate as black-box models with limited interpretability, making it difficult to derive actionable insights or support real-time operational decision-making. Furthermore, these methods are not designed to interoperate with other critical tasks such as anomaly detection or trajectory forecasting. These limitations highlight the need for a new framework that not only improves predictive accuracy but also enables context-aware reasoning and seamless integration across multiple maritime analytics tasks.

\subsection{Maritime Anomaly Detection}
Maritime anomaly detection based on AIS data plays a critical role in enhancing maritime situational awareness, aiming to identify abnormal vessel behaviors such as abrupt course changes, route deviations, or AIS signal interruptions. GeoTrackNet~\cite{nguyen2021geotracknet} employs a variational recurrent neural network (VRNN) to learn probabilistic representations of AIS trajectories and utilizes a geospatially-aware a-contrario detection framework to identify anomalous patterns. GTRA~\cite{singh2022leveraging} introduces a graph-based traffic representation combined with evidential deep learning to capture both trajectory uncertainties and anomaly likelihoods. These models have demonstrated promising performance in learning complex spatiotemporal patterns and detecting diverse types of anomalies.

However, a critical limitation of most existing approaches is their lack of interpretability. While they effectively determine whether a given behavior is anomalous, they typically do not provide semantically meaningful explanations for why the anomaly occurred or how it is situated within the broader maritime context. This impedes their applicability in operational settings, where actionable insights and human-interpretable justifications are essential for decision-making and response planning. To address this gap, there is a growing demand for anomaly detection frameworks that offer both accurate identification and context-aware, explainable reasoning.

\subsection{Time Series Forecasting}
Time series forecasting plays a pivotal role in domains such as energy demand, traffic flow, and weather prediction. Recent Transformer-based models—e.g., Autoformer~\cite{wu2021autoformer}, TimesNet~\cite{wu2023timesnet}, and iTransformer~\cite{liu2023itransformer}—have advanced long-term forecasting by effectively modeling seasonality, trends, and inter-variable interactions. However, these models are generally designed for single-task settings with domain-specific objectives, limiting their applicability to complex environments like maritime traffic systems.

Most prior works forecast a single variable for an individual vessel, neglecting inter-vessel correlations and cross-task dependencies. As a result, shared behavioral patterns across vessels and interactions among variables such as position, speed, and direction are often overlooked. Furthermore, isolated task-specific learning hinders knowledge transfer across forecasting objectives.

To address these limitations, we propose a unified framework that jointly models multiple vessels and tasks, including trajectory prediction, anomaly detection, and collision risk assessment. By integrating sequential AIS data with large language models (LLMs), our AIS-LLM framework captures contextual dependencies, supports cross-task reasoning, and generates explainable, semantically enriched outputs for maritime decision-making.


\begin{figure*}[t]
    \centering
    \includegraphics[width=0.85\linewidth]{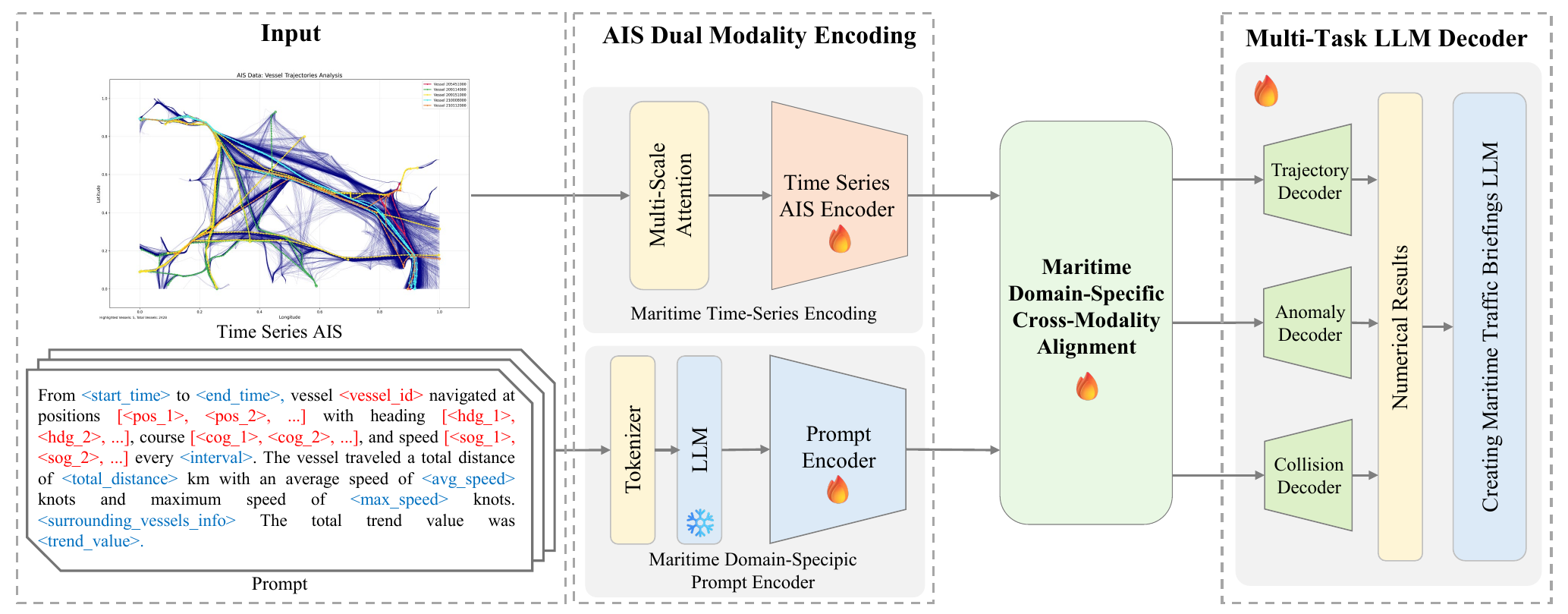}
    \caption{\textbf{Overview of the AIS-LLM framework.} From left to right, (1) AIS time series data and its input converted into text prompts, (2) a bi-modality encoding module that includes a time series encoder and a prompt encoder, (3) information integration via cross-modality attentions, and (4) a multi-task LLM decoder that performs trajectory prediction, anomaly detection, collision risk assessment, and time series prediction. This integrated architecture provides both numerical prediction and natural language interpretation simultaneously, enabling comprehensive analysis of maritime traffic conditions.}
    \label{fig:ais_llm_pdf}
\end{figure*}

\section{Method}
\subsection{System Overview}
Figure~\ref{fig:ais_llm_pdf} illustrates the AIS-LLM framework designed to analyze AIS signals. AIS data includes timestamps, geographic coordinates (longitude, latitude), vessel identifiers, heading, course, rate of turn, and speed. It presents several challenges due to the non-linear behavioral patterns of vessels, multivariate dependencies, and complex inter-vessel interactions.

To address these challenges, AIS-LLM integrates AIS time-series data with a large language model (LLM)~\cite{team2024qwen2} and is composed of the following four core components: (1) a multi-scale time-series encoder to process AIS sequences, (2) a prompt encoder that converts structured AIS data into natural language, (3) a cross-modal alignment module for semantic integration of embeddings, and (4) a multi-task decoder that generates both numerical predictions and interpretable text explanations. The multi-task decoder simultaneously performs trajectory prediction, anomaly detection, and collision risk assessment.

\subsection{Multi-Scale Time-Series Encoding}
The AIS encoder incorporates Multi-Scale Temporal Reasoning to effectively process maritime situations across different time scales. Maritime traffic involves a wide range of temporal resolutions---from immediate collision risks to long-term prediction. Fixed time window approaches struggle to capture this variability.

Inverted embedding condenses temporal AIS information (e.g., positional and kinematic data) into a single token embedding using a Pre-Layer Normalization (Pre-LN) Transformer~\cite{huang2023normalization}. Unlike conventional methods that encode each timestamp individually, this approach encodes the entire time-series of each feature as a single token, thus capturing long-term dependencies more effectively:
\begin{equation}
\mathbf{H}^T = \text{TSEncoder}(\mathbf{W}_e \mathbf{X} + \mathbf{b}_e)
\end{equation}
Here, \( \mathbf{X} \in \mathbb{R}^{T \times 4} \) denotes the AIS time-series data (e.g., latitude and longitude), and TSEncoder is a Pre-LN Transformer encoder.

Multi-scale temporal attention allows each attention head to handle a different time resolution (scales = [1, 4, 16, 32]), enabling hierarchical processing from short-term responses to long-term prediction:
\begin{equation}
\mathbf{H}_{\text{multi}} = \text{MultiScaleAttention}(\mathbf{H}^T, \text{scales} = [1, 4, 16, 32])
\end{equation}
This enables the LLM to perform complex reasoning such as: ``There is a high risk of collision in the short term, but the long-term route is safe.''

To normalize time-series variables, Reversible Instance Normalization (RevIN) is applied per AIS variable, and the original scale is restored after prediction. This is crucial for handling AIS variables that differ in units and range (e.g., latitude/longitude in degrees, speed in knots, heading in degrees).

\subsection{LLM-based Prompt Encoding}
\label{prompt_encoding}
AIS data is converted into natural language prompts to make it compatible with the LLM (Qwen2-1.5B~\cite{team2024qwen2}). These prompts concisely summarize vessel trajectories, including key positional data, motion characteristics, temporal features, and contextual information. The final token embedding from the LLM encoder is used as the semantic representation and is further refined using a lightweight Transformer layer to enhance maritime domain specificity.

\subsection{Cross-Modal Alignment}
To effectively integrate the time-series and textual embeddings, a selective cross-attention mechanism is introduced. This mechanism aligns detailed vessel movement patterns captured by the AIS embeddings with high-level maritime domain knowledge encoded in the LLM embeddings. The aligned representations adaptively reflect maritime regulations and geographic constraints, improving performance in trajectory prediction, anomaly detection sensitivity, and collision risk evaluation.

\subsection{Multi-Task LLM Decoder}
\label{llm_decoder}
The multi-task decoder simultaneously performs vessel trajectory prediction, anomaly detection, and collision risk evaluation. Each task-specific module is based on a Transformer architecture and shares the aligned, domain-integrated embeddings. Each task’s numerical result is complemented with a textual explanation generated by the LLM (Qwen2-1.5B\cite{team2024qwen2}).

To address the issue of limited data, AIS-LLM introduces a self-supervised learning mechanism that leverages consistency between generated textual explanations and numerical predictions. During the explanation generation training phase, teacher forcing is applied to facilitate efficient learning. Specifically, both the prompt (including numerical prediction results) and the ground-truth explanation are fed into the LLM decoder, which is trained to predict the next token. During inference, the decoder generates maritime traffic briefings in an autoregressive manner based solely on the numerical prediction results.

Training is guided by a composite loss function consisting of trajectory prediction (MAE), anomaly detection (BCE loss), collision risk quantification (MSE), and explanation generation. Task-specific weights are applied to balance performance across tasks, ensuring comprehensive and interpretable maritime traffic analysis.

For detailed implementation, please refer to Appendix.



\section{Experiment}

\begin{table*}[ht]
\centering
\small
\begin{tabular}{lccccccccccccccc}
\toprule
\multirow{2}{*}{Models} & \multicolumn{2}{c}{\textbf{AIS-LLM}} & \multicolumn{2}{c}{TrAISformer} & \multicolumn{2}{c}{iTransformer} & \multicolumn{2}{c}{iFlowformer} & \multicolumn{2}{c}{iFlashformer} & \multicolumn{2}{c}{iInformer} & \multicolumn{2}{c}{iReformer} \\
& \multicolumn{2}{c}{\textbf{(Ours)}} & \multicolumn{2}{c}{(2024)} & \multicolumn{2}{c}{(2024)} & \multicolumn{2}{c}{(2024)} & \multicolumn{2}{c}{(2024)} & \multicolumn{2}{c}{(2024)} & \multicolumn{2}{c}{(2024)} \\
\cmidrule(lr){2-3} \cmidrule(lr){4-5} \cmidrule(lr){6-7} \cmidrule(lr){8-9} \cmidrule(lr){10-11} \cmidrule(lr){12-13} \cmidrule(lr){14-15}
Metric & ADE & FDE & ADE & FDE & ADE & FDE & ADE & FDE & ADE & FDE & ADE & FDE & ADE & FDE \\
\midrule
Piraeus & \textbf{0.43} & \textbf{0.91} & 0.66 & 1.22 & 0.50 & 1.10 & 0.50 & 1.10 & 0.50 & 1.11 & 0.48 & 1.05 & 0.52 & 1.12 \\
DMA & \textbf{3.11} & \textbf{4.98} & 3.30 & 7.91 & 7.08 & 15.27 & 6.72 & 14.55 & 6.87 & 14.90 & 6.20 & 13.10 & 6.79 & 14.85 \\
\bottomrule
\end{tabular}
\caption{Performance comparison of vessel trajectory prediction (nautical mile)}
\label{tab:trajectory_prediction}
\end{table*}

\subsection{Dataset}
This study utilizes two AIS datasets collected from distinct maritime regions and periods to support the analysis of vessel traffic patterns. Together, they offer complementary perspectives on maritime dynamics.

\subsubsection{The Piraeus AIS Dataset}
The Piraeus AIS dataset~\cite{tritsarolis2022piraeus}, provided by the University of Piraeus, contains vessel navigation data collected from May 2017 to December 2019. The Region of Interest (ROI) spans the Saronic Gulf near the Port of Piraeus, specifically bounded between (37.5°, 23.1°) and (37.9°, 23.7°).

To enhance data quality, AIS messages were filtered to remove unrealistic speeds over 30 knots, moored vessels, and messages within 1 nautical mile of the coastline. Trajectories were segmented when time gaps exceeded 30 minutes, resampled at 1-minute intervals, and linearly interpolated. Only segments with at least 42 AIS messages were retained.

The final dataset comprises 1,800 trajectories for training and 451 for testing. A full description of the preprocessing pipeline is provided in Appendix.

\subsubsection{The Danish Maritime Authority Dataset}
The DMA AIS dataset~\cite{ray2019heterogeneous}, released by the Danish Maritime Authority, contains AIS messages collected throughout 2019 for cargo and tanker vessels navigating in Danish waters. The ROI covers a maritime area between (55.5°, 10.3°) and (58.0°, 13.0°).

Data preprocessing followed the methodology proposed in ~\citet{nguyen2024transformer}, removing AIS messages with speeds above 30 knots, moored or anchored statuses, and those within 1 nautical mile of shore. Trajectories were segmented when time gaps exceeded two hours, and any segments with fewer than 20 messages or lasting less than 4 hours were discarded. Data were then resampled at uniform 10-minute intervals, and voyages longer than 20 hours were split.

After preprocessing, the dataset includes 8,462 training trajectories and 1,397 test trajectories. Additional preprocessing details are presented in Appendix.

\subsection{Implementation Details}
\label{'implementation details'}
All experiments were conducted using four NVIDIA RTX A6000 GPUs. Both the prompt encoder and the LLM decoder were based on the Qwen2-1.5B model, with the LLM decoder fine-tuned using QLoRA for parameter-efficient adaptation. To optimize memory usage, 8-bit quantization was applied. For detailed implementation settings, please refer to the Appendix.

\begin{table}[ht]
\centering
\small
\begin{tabular}{lccc}
\toprule
Models & Precision & Recall & F1-Score \\
\midrule
TimesNet (2023)         & 0.65 & 0.32 & 0.43 \\
Autoformer (2021)       & 0.59 & 0.40 & 0.48 \\
iTransformer (2024)     & 0.58 & 0.34 & 0.43 \\
DLinear (2023)          & 0.65 & 0.33 & 0.44 \\
\textbf{AIS-LLM (Ours)} & \textbf{0.66} & \textbf{0.44} & \textbf{0.53} \\
\bottomrule
\end{tabular}
\caption{Performance comparison of maritime anomaly detection models on the Piraeus dataset.}
\label{tab:anomaly_detection}
\end{table}

\begin{table}[ht]
\centering
\small
\begin{tabular}{lcc}
\toprule
Model & MAE & RMSE \\
\midrule
iTransformer (2024)     & 0.0460 & 0.0752 \\
iFlowformer (2024)      & 0.0469 & 0.0764 \\
iFlashformer (2024)     & 0.0461 & 0.0755 \\
iReformer (2024)        & 0.0466 & 0.0761 \\
\textbf{AIS-LLM (Ours)} & \textbf{0.0414} & \textbf{0.0610} \\
\bottomrule
\end{tabular}
\caption{Performance comparison of vessel collision risk assessment models on the DMA dataset.}
\label{tab:collion}
\end{table}

\begin{table*}[ht]
\centering
\small
\begin{tabular}{lcccccccccccccc}
\toprule
\multirow{2}{*}{Models} & \multicolumn{2}{c}{\textbf{AIS-LLM}} & \multicolumn{2}{c}{TrAISformer} & \multicolumn{2}{c}{iTransformer} & \multicolumn{2}{c}{iFlowformer} & \multicolumn{2}{c}{iFlashformer} & \multicolumn{2}{c}{iInformer} & \multicolumn{2}{c}{iReformer} \\
& \multicolumn{2}{c}{\textbf{(Ours)}} & \multicolumn{2}{c}{(2024)} & \multicolumn{2}{c}{(2024)} & \multicolumn{2}{c}{(2024)} & \multicolumn{2}{c}{(2024)} & \multicolumn{2}{c}{(2024)} & \multicolumn{2}{c}{(2024)} \\
\cmidrule(lr){2-3} \cmidrule(lr){4-5} \cmidrule(lr){6-7} \cmidrule(lr){8-9} \cmidrule(lr){10-11} \cmidrule(lr){12-13} \cmidrule(lr){14-15}
Metric & MSE & MAE & MSE & MAE & MSE & MAE & MSE & MAE & MSE & MAE & MSE & MAE & MSE & MAE \\
\midrule
Piraeus & \textbf{95.76} & \textbf{1.96} & 268.98 & 5.13 & 272.16 & 5.13 & 279.42 & 5.26 & 275.86 & 5.20 & 259.52 & 4.92 & 278.55 & 5.30 \\
DMA & \textbf{111.44} & \textbf{3.00} & 341.90 & 4.78 & 561.48 & 8.00 & 576.58 & 8.24 & 569.65 & 8.07 & 581.34 & 8.00 & 598.92 & 8.41 \\
\bottomrule
\end{tabular}
\caption{Performance comparison of time series forecasting}
\label{tab:time_series_forecasting}
\end{table*}


\begin{table}[ht]
\centering
\small
\begin{tabular}{lccc}
\toprule
Model & BLEU-4 & ROUGE-L & BERTScore \\
\midrule
Qwen2-1.5B*                & 0.231 & 0.267 & 0.401 \\
LLaMA-3-8B                 & 0.211 & 0.289 & 0.517 \\
Mistral-7B-Instruct       & 0.198 & 0.264 & 0.534 \\
Qwen3-8B                    & 0.267 & 0.325 & 0.612 \\
\textbf{AIS-LLM (Ours)}     & \textbf{0.412} & \textbf{0.486} & \textbf{0.666} \\
\bottomrule
\end{tabular}
\caption{Performance comparison of text generation models on the Piraeus dataset. *Denotes pre-finetuning baseline model.}
\label{tab:model_comparison}
\end{table}

\begin{figure*}[t]
    \centering
    \includegraphics[width=0.9\linewidth]{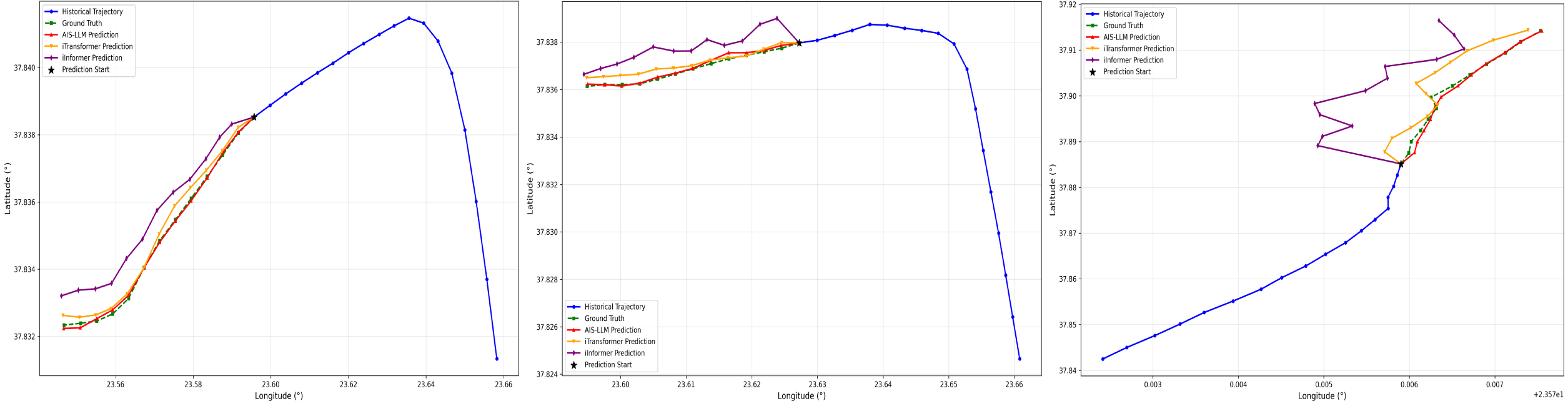}
    \caption{\textbf{Trajectory Prediction Performance Comparison on Piraeus.} Qualitative comparison of vessel trajectory prediction results between the proposed AIS-LLM framework and baseline models. Historical Trajectory (blue solid line) denotes the observed past trajectory, while Ground Truth (green dashed line) represents the future trajectory to be predicted. Predicted trajectories are shown for AIS-LLM (red), iTransformer (yellow), and iInformer (purple), respectively.}
    \label{fig:trajectory_prediction_result}
\end{figure*}

\begin{figure*}[t] 
    \centering
    \includegraphics[width=0.85\linewidth]{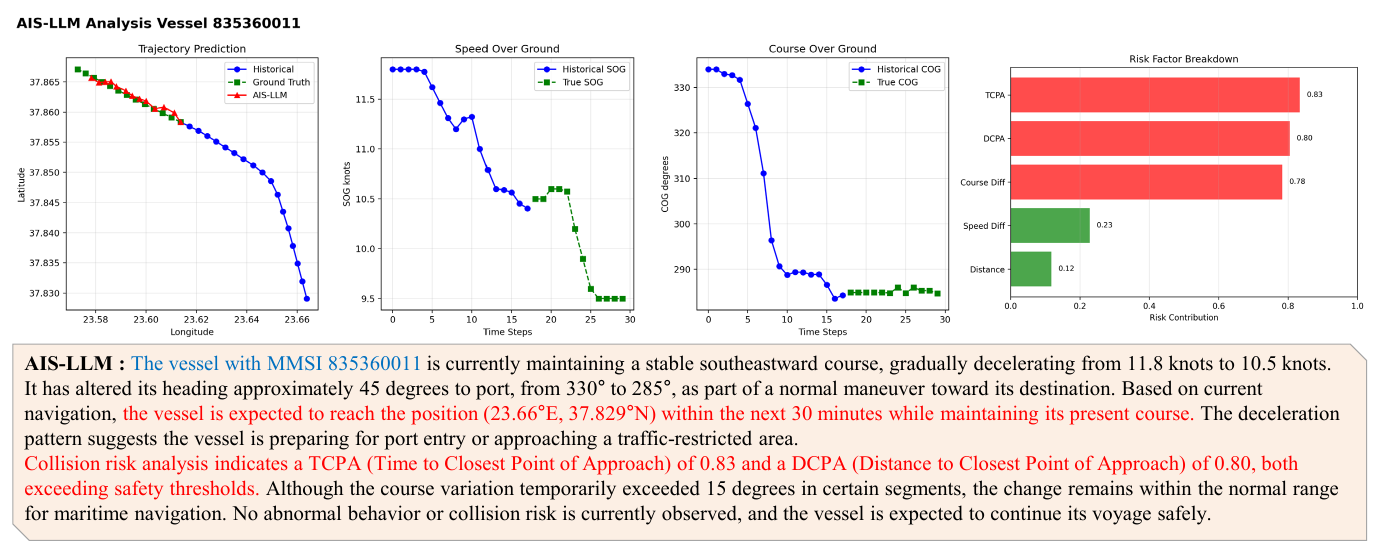}
    \caption{\textbf{Multimodal natural language outputs generated by AIS-LLM.} AIS-LLM analyzes maritime traffic conditions using vessel data and produces interpretable natural language descriptions. It explains trajectory predictions with insights on expected movement and speed variations, identifies anomalies with descriptions of abnormal behaviors and severity, and assesses collision risks by interpreting CPA details. All outputs combine quantitative analysis with human-readable explanations via the LLM decoder, offering intuitive support for maritime decision-making.}
    \label{fig:briefing_prediction}
\end{figure*}

\subsection{Quantitative Evaluation Results}
\subsubsection{Vessel Trajectory Prediction}
Table \ref{tab:trajectory_prediction} presents a quantitative comparison of vessel trajectory prediction performance, measured in nautical miles, across two maritime regions: the Port of Piraeus and the Danish Maritime Area (DMA). The Piraeus dataset consists of AIS records sampled at 1-minute intervals, while the DMA dataset is sampled at 10-minute intervals. All models were tasked with forecasting 24 future positions based on 18 historical observations. Evaluation was conducted using two standard metrics: Average Displacement Error (ADE) and Final Displacement Error (FDE).

In the Piraeus setting, which requires fine-grained, short-range predictions, all models achieved relatively low error rates. The proposed AIS-LLM model attained the best performance, with an ADE of 0.43 and an FDE of 0.91. These results represent a performance improvement over the previous best-performing model, iInformer, which reported an ADE of 0.48 and an FDE of 1.05. This demonstrates AIS-LLM’s capability to effectively capture short-term vessel dynamics in high-resolution temporal settings.

\subsubsection{Maritime Anomaly Detection}
Table \ref{tab:anomaly_detection} compares maritime anomaly detection models applied to the Port of Piraeus, evaluating the proposed AIS-LLM against recent deep learning models using Precision (P), Recall (R), and F1 score metrics. Anomaly data was generated following the ~\citet{li2024stad} methodology.

AIS-LLM achieved superior performance across all metrics (Precision: 0.66, Recall: 0.44, F1-score: 0.53), outperforming previous leading models including Autoformer (Precision: 0.59, F1-score: 0.48) and DLiner (F1-score: 0.44). Other models like TimesNet and iTransformer showed high Precision but low Recall, resulting in lower F1-scores. These results demonstrate AIS-LLM's capability to detect various irregular movement patterns while maintaining robust performance in complex, high-density maritime environments.

\subsubsection{Vessel Collision Risk Assessment}

Table~\ref{tab:collion} presents the collision risk prediction performance of AIS-LLM compared to several recent deep time-series forecasting models. The evaluation was conducted on the DMA dataset following the protocol and metrics (MAE, RMSE) defined in \citet{tritsarolis2023collision}.

AIS-LLM achieved the best performance across both metrics, with a MAE of 0.0414 and an RMSE of 0.0610, significantly outperforming all baselines. These results demonstrate AIS-LLM’s enhanced capability to estimate the Collision Risk Index (CRI) with high precision. In safety-critical scenarios such as maritime collision risk assessment, reducing RMSE is particularly crucial, as it lowers the chance of underestimating dangerous situations. The findings underscore AIS-LLM’s potential as a reliable decision-support tool for real-time vessel traffic monitoring systems.

\subsubsection{Time Series Forecasting}
Table \ref{tab:time_series_forecasting} compares time series forecasting performance across the Port of Piraeus and Danish Maritime Area (DMA). All models forecast 24 future time steps based on 18 historical observations, evaluated using MSE and MAE metrics.

AIS-LLM consistently outperformed all baselines. For Piraeus' high-resolution 1-minute AIS data, AIS-LLM achieved the lowest errors (MSE: 95.76, MAE: 2.11), significantly improving upon TrAISformer (MSE: 268.98, MAE: 5.13) and iTransformer (MSE: 272.16, MAE: 5.13). In the more complex DMA dataset with lower temporal resolution, AIS-LLM maintained superior performance (MSE: 111.44, MAE: 3.00) while competing models exceeded MSE values of 500. These results validate AIS-LLM's robustness across diverse maritime scenarios and temporal contexts.

\subsubsection{Natural Language Generation Performance}

Table~\ref{tab:model_comparison} reports the natural language generation performance of various models~\cite{team2024qwen2,dubey2024llama,jiang2024mixtral,yang2025qwen3} evaluated on the Piraeus dataset. Since there is no existing benchmark dataset tailored to the maritime domain, we constructed a custom evaluation set through hard-coded annotations specific to maritime scenarios. 

AIS-LLM shows the best performance across all metrics, achieving the highest scores in BLEU-4, ROUGE-L, and BERTScore. These results indicate that AIS-LLM is effective in generating text that is both lexically and semantically aligned with expert annotations. Compared to strong baselines such as Qwen3-8B and LLaMA-3-8B, AIS-LLM demonstrates clear advantages in fluency and contextual relevance.

\subsection{Qualitative Evaluation Results}
To validate the practical effectiveness of AIS-LLM, we qualitatively analyzed real-world vessel trajectories from the DMA dataset. Figures~\ref{fig:trajectory_prediction_result} and~\ref{fig:briefing_prediction} illustrate the model’s capabilities in trajectory prediction, collision risk assessment, and multimodal explanation.


Figure~\ref{fig:trajectory_prediction_result} presents a qualitative comparison of vessel trajectory predictions generated by AIS-LLM and two representative baseline models, iTransformer and iInformer, on the Piraeus dataset. The figure illustrates both the observed historical trajectories (blue solid line) and the ground truth future trajectories (green dashed line), alongside the predicted outputs from each model.

AIS-LLM's predicted trajectories (red) align more closely with the ground truth paths than those of the baseline models, particularly in capturing directional changes and subtle turns. In contrast, iTransformer (yellow) and iInformer (purple) often exhibit oversmoothing or deviation in high-curvature regions, leading to reduced prediction fidelity. These results visually support the quantitative gains reported in earlier sections, highlighting AIS-LLM’s superior ability to model vessel dynamics in complex maritime environments.



Figure~\ref{fig:briefing_prediction} highlights the interpretability capabilities of AIS-LLM through multimodal natural language outputs generated from vessel trajectory and navigation data. The framework integrates time-series AIS inputs to produce coherent textual explanations that accompany quantitative visualizations.

The figure illustrates how AIS-LLM generates context-aware narratives, including trajectory forecasts, anticipated waypoints, and course changes, while also detecting potential anomalies and assessing collision risk based on TCPA and DCPA metrics. The generated description not only identifies threshold violations in safety-critical parameters but also interprets their navigational implications in plain language. This level of interpretability enhances situational awareness and supports human operators in making informed maritime navigation decisions.



\section{Conclusion}
In this paper, we introduced AIS-LLM, a unified framework that integrates time-series AIS data with large language models to simultaneously perform vessel trajectory prediction, anomaly detection, and collision risk assessment. By leveraging a multi-scale time-series encoder, a maritime-specific prompt encoder, and a cross-modality alignment module, AIS-LLM delivers not only accurate numerical forecasts but also context-aware natural language explanations. This dual-modality design enhances both the interpretability and operational value of model outputs, thereby supporting informed maritime decision-making.

Extensive experiments conducted on two real-world AIS datasets demonstrate that AIS-LLM consistently outperforms state-of-the-art baselines across all tasks. The framework achieves superior accuracy in trajectory prediction and collision risk estimation while maintaining robust performance in detecting anomalous vessel behaviors. Moreover, AIS-LLM exhibits strong capabilities in generating coherent, semantically-rich textual summaries, outperforming competitive language models on maritime-specific generation tasks.

The results validate the effectiveness of our multi-task learning approach in capturing inter-task dependencies and vessel interactions, offering a scalable and interpretable solution for complex maritime traffic analysis. Future research may extend AIS-LLM by incorporating environmental factors (e.g., weather, ocean currents) and exploring real-time deployment scenarios for proactive maritime safety and navigation support.


\bibliography{aaai2026}


\clearpage
\appendix

\begin{table*}[t]
\centering
\footnotesize
\begin{tabular}{lcccccc}
\toprule
Dataset & Latitude Range & Longitude Range & Train Waypoints & Test Waypoints & Train Trajectories & Test Trajectories \\
\midrule
Piraeus & 37.5°--37.9° & 23.1°--23.7° & 93,332 & 23,524 & 1,800 & 451 \\
DMA & 55.5°--58.0° & 10.3°--13.0° & 690,761 & 119,525 & 8,462 & 1,397 \\
\bottomrule
\end{tabular}
\caption{Statistics of the vessel trajectory datasets used in our experiments.}
\label{tab:dataset_statistics}
\end{table*}

\section{A. Dataset Details and Preprocessing}
\subsection{A.1 AIS Datasets Overview}
We utilize two AIS datasets collected from distinct maritime regions to ensure diversity in geographical and navigational patterns. Table~\ref{tab:dataset_statistics} summarizes their spatial coverage and the number of extracted waypoints and vessel trajectories used in our study. The datasets differ in scope and scale, providing complementary perspectives for robust model evaluation.

\subsubsection{A.1.1 The Piraeus AIS Dataset}
The Piraeus AIS dataset~\cite{tritsarolis2022piraeus} comprises vessel trajectory data collected over a period of 2.5 years (May 2017 to December 2019) from ships navigating in the vicinity of the Port of Piraeus, one of the busiest ports in Europe. Data acquisition was performed via a terrestrial AIS receiver located at the University of Piraeus. The dataset includes both dynamic (kinematic) and static vessel information, with each message timestamped and geo-referenced using WGS-84 coordinates. The coverage spans approximately 10,000 square kilometers, encompassing the Saronic Gulf region.

This high-density dataset captures a wide variety of vessel activities, including commercial shipping, passenger transport, and leisure traffic. The distribution of vessel types is notably diverse, with a significant presence of cargo ships, tankers, and passenger vessels, reflecting the strategic importance of the Piraeus port as a regional maritime hub. This diversity makes the dataset particularly suitable for training and evaluating models under realistic, heterogeneous traffic conditions.

\subsubsection{A.1.2 The DMA AIS Dataset}
The DMA AIS dataset~\cite{ray2019heterogeneous} is a heterogeneous maritime dataset covering a wide region that includes the Celtic Sea, the English Channel, the North Atlantic Ocean, and the Bay of Biscay. It contains AIS messages recorded over a six-month period (October 2015 to March 2016) via a terrestrial receiver operated by the French Naval Academy. The dataset integrates multiple data categories: dynamic and static AIS messages, vessel registries, geographic layers, and environmental metadata.

A key characteristic of this dataset is the rich diversity of vessel types, particularly fishing vessels, cargo ships, and service/support vessels operating in offshore and coastal waters. The AIS data is complemented by vessel registry databases, enabling a more detailed classification of ship types. This characteristic enhances the dataset's utility for analyzing complex maritime behaviors and supports applications in traffic modeling, anomaly detection, and multi-class trajectory learning tasks.

\subsubsection{A.1.3 Exclusion of Geospatial and External Domain-Specific Features}
In this study, we intentionally exclude external domain-specific features such as meteorological conditions, coastline proximity, maritime protected areas, and predefined shipping lanes from the model input. Instead, we rely exclusively on fundamental navigational variables derived from AIS messages—specifically, vessel position (latitude and longitude), speed over ground (SOG), and course over ground (COG).

This design choice is guided by two key considerations. First, by constraining the input space to core navigational features, we aim to enable a more transparent assessment of model performance, where improvements can be directly attributed to architectural design rather than the inclusion of auxiliary data. This facilitates fairer benchmarking and clearer interpretation of results. Second, such a minimalist input setting enhances the model's generalizability across diverse maritime environments, particularly in scenarios where domain-specific contextual data may be unavailable, incomplete, or unreliable.

Accordingly, our approach emphasizes architectural robustness and domain-agnostic applicability over reliance on region-specific features, promoting broader applicability and reproducibility.

\subsection{A.2 Preprocessing Pipeline}
\subsubsection{A.2.1 The Piraeus AIS Dataset}
To enhance the quality and reliability of AIS data, a series of preprocessing steps were applied to the Piraeus dataset. First, AIS messages with unrealistic speed values (Speed Over Ground, SOG \textgreater 30 knots) were removed, as such values are likely to result from sensor or transmission errors. Messages from vessels identified as moored or at anchor were also excluded, in order to focus solely on dynamic navigation behaviors. Furthermore, any AIS points located within 1 nautical mile of the coastline were discarded to minimize the influence of harbor operations and coastal noise.

Voyage trajectories were segmented when the time gap between consecutive AIS messages exceeded 305 minutes, ensuring temporal continuity within each segment. Linear interpolation was applied to resample all trajectories at uniform 1-minute intervals, thereby achieving consistent temporal resolution across the dataset. Segments with fewer than 42 AIS points were excluded, as they were deemed insufficient for robust model training. To control the computational complexity and memory usage during training, long trajectories were segmented into shorter overlapping sequences using a sliding window approach with a maximum sequence length of 150 AIS points.

Finally, all input variables were normalized using Min-Max scaling to the range [0, 1], facilitating stable training and convergence of the learning models.

\subsubsection{A.2.2 The DMA AIS Dataset}
The preprocessing of the DMA dataset largely followed the procedure outlined in \citet{nguyen2024transformer}, with additional refinements tailored to our model’s input requirements. AIS messages indicating moored or anchored statuses were removed, along with those reporting speeds greater than 30 knots or located within 1 nautical mile of the shoreline. To ensure temporal coherence, trajectories were split when the interval between consecutive messages exceeded two hours. Voyage segments were retained only if they contained at least 20 AIS messages and spanned a minimum of four hours in duration.

To enforce consistency in data representation, trajectories were resampled at fixed 10-minute intervals using linear interpolation. Segments exceeding 20 hours in length were divided into smaller, manageable subsequences.  A minimum trajectory length threshold of 42 AIS points was imposed to ensure each segment provided sufficient temporal context for downstream tasks.

As with the Piraeus dataset, all numerical input features were normalized using Min-Max scaling.

\section{B. Implementation Details}
\subsection{B.1 Model Architecture Specifications}

\subsubsection{B.1.1 Detailed Structure of Time Series Encoder}
The time-series encoder in our framework is inspired by the Inverted Embedding approach introduced in TimeCMA, specifically designed to effectively process the unique characteristics of AIS data. AIS data consist of heterogeneous variables with distinct physical meanings and scales, such as position (latitude, longitude), speed over ground (SOG), and course over ground (COG). Each variable exhibits independent temporal patterns. For instance, latitude and longitude form the spatial trajectory of a vessel, whereas speed and course represent the vessel’s dynamic state.

\paragraph{Inverted Embedding Layer.}
The input time-series $X \in \mathbb{R}^{B \times T \times N}$ is transposed to $X \in \mathbb{R}^{B \times N \times T}$, treating each AIS variable as an independent channel. This enables the model to learn variable-specific temporal dependencies, effectively capturing the continuity of position variables and the fluctuation patterns of speed and course. A linear projection 
\[
f_{\text{embed}}(X) = X \times W_{\text{embed}} + b_{\text{embed}}
\]
is applied to transform the temporal dimension $T$ into the feature dimension $d_{\text{model}} = 512$, compressing each variable's temporal sequence into a high-dimensional feature vector.

\paragraph{Pre-Layer Normalization Transformer Encoder.}
To model the complex dynamics of maritime environments, we use $L=2$ encoder layers. Each layer is defined as:
\begin{align*}
    \text{Encoder}_\ell(H) &= \text{LayerNorm}(H + \text{MHA}(H)), \\
    \text{Encoder}_{\ell+1}(H') &= \text{LayerNorm}(H' + \text{FFN}(H')),
\end{align*}
where MHA denotes Multi-Head Attention and FFN denotes a Feed-Forward Network. Pre-Layer Normalization ensures stable training in deep networks, providing robustness against irregular sampling intervals and noise commonly found in AIS data. Through multi-head attention, the encoder effectively captures complex inter-variable interactions.
\subsubsection{B.1.2 Multi-Scale Temporal Attention}
In maritime environments, decision-making spans multiple time scales, from immediate collision avoidance to long-term route planning. To capture this characteristic, we design a \textbf{Multi-Scale Temporal Attention} mechanism.

\paragraph{Scale-Specific Attention Heads.}
Each temporal scale $s \in \{1, 4, 16, 32\}$ corresponds to a different time horizon of maritime decision-making. Specifically:
\begin{itemize}
    \item \textbf{Scale 1} models immediate responses in emergency situations,
    \item \textbf{Scale 4} captures short-term maneuvering plans,
    \item \textbf{Scale 16} reflects mid-term route adjustments,
    \item \textbf{Scale 32} encodes long-term strategic planning.
\end{itemize}
For each scale $s$, attention is computed as:
\[
\text{Attention}_s(Q, K, V) = \text{softmax} \left( \frac{QK^\top}{\sqrt{d_k}} \cdot \alpha_s \right) V,
\]
where $\alpha_s$ is a learnable parameter that modulates temporal sensitivity, allowing the model to focus on the most relevant time scale depending on the context.

\paragraph{Scale-Specific Positional Encoding.}
To effectively learn periodic temporal patterns, we apply a scale-specific positional encoding:
\[
\text{PE}_s(\text{pos}, 2i) = \sin\left( \frac{\text{pos}}{10000^{2i/d_{\text{model}}}} \cdot s \right),
\]
which adjusts the frequency of the encoding to match each temporal resolution.

\paragraph{Cross-Scale Fusion.}
To integrate information across different time scales, we apply cross-scale attention over concatenated features from each scale:
\[
F_{\text{cross}} = \text{Attention} \left( \text{concat}(F_1, F_4, F_{16}, F_{32}) \right),
\]
\[
F_{\text{fused}} = \text{Linear}(F_{\text{cross}}) + \text{Residual}(H_{\text{input}}),
\]
where $F_s$ denotes the output of the attention head at scale $s$. This fusion allows the model to dynamically combine insights from multiple temporal perspectives.
\subsubsection{B.1.3 Cross-Modality Alignment Module}
To effectively combine the numerical characteristics of AIS data with the expressive power of large language models (LLMs), we design a Cross-Modality Alignment module. While a pure time-series encoder excels at capturing disentangled temporal patterns, it can be limited in expressive capacity. To address this, we incorporate LLM-based encoding as a complementary representation pathway.
\paragraph{Alignment Mechanism.}
We apply cross-attention where the time-series embedding \( H_{\text{ts}} \) serves as the query, and the LLM embedding \( H_{\text{llm}} \) serves as both key and value:
\[
H_{\text{aligned}} = \text{Attention}(H_{\text{ts}}, H_{\text{llm}}, H_{\text{llm}}).
\]
This alignment bridges the gap between the precise but less expressive time-series representation and the rich, robust prompt embeddings generated by the LLM. The LLM branch converts AIS data into natural language templates, producing semantically enriched representations that go beyond raw numerical patterns and enable more robust feature extraction.

\paragraph{Dimension Alignment.}
To match the dimensionality of the LLM embeddings (\( d_{\text{llm}} = 1536 \)) with the number of AIS variables (\( N = 4 \)), we apply a linear projection:
\[
H_{\text{llm\_proj}} = \text{Linear}_{d_{\text{llm}} \rightarrow N}(H_{\text{llm}}).
\]
This maps the high-dimensional robust prompt embeddings into the physical variable space of AIS data (latitude, longitude, SOG, COG), enabling effective cross-modal alignment within a shared representational space.

\subsubsection{B.1.4 LLM Decoder and QLoRA Application}
To enable explainable maritime traffic analysis, we construct a language model decoder based on the Qwen2-1.7B model. Since AIS data involve complex numerical changes in multiple variables over time, translating these patterns into natural language enables richer contextual and detailed information delivery. This helps users to clearly understand maritime situations beyond what raw numerical summaries can convey.

\paragraph{Necessity of QLoRA.}
Fully fine-tuning a large language model is computationally expensive and susceptible to overfitting. To address this, we adopt QLoRA (Quantized Low-Rank Adaptation) to enable efficient and lightweight adaptation. The weight update is defined as:
\[
W = W_0 + \alpha BA,
\]
where \( W_0 \) denotes the frozen pretrained weights, and only the low-rank matrices \( B \in \mathbb{R}^{d \times r} \) and \( A \in \mathbb{R}^{r \times k} \) are trained. This approach injects domain-specific knowledge into the model while updating only about 0.1\% of the total parameters, achieving effective domain adaptation with minimal computational overhead.

\paragraph{Task-Specific Output Heads.}
To support multiple downstream tasks, we design specialized output heads tailored to each task. For trajectory prediction, we apply a linear transformation \( \text{Linear}(d_{\text{model}} \rightarrow \text{pred}_{\text{len}} \times n_{\text{vars}}) \), which predicts future positions and motion states. For anomaly detection, we use a linear layer \( \text{Linear}(d_{\text{model}} \rightarrow 2) \) followed by a Softmax function to classify normal versus anomalous behaviors. For collision risk estimation, we adopt a linear layer \( \text{Linear}(d_{\text{model}} \rightarrow 1) \) followed by a Sigmoid function to output a collision risk index (CRI) in the range of [0, 1].

\subsection{B.2 Hyperparameter Settings and Optimization}
To train the proposed framework, we apply the Cosine Annealing with Warm Restarts learning rate scheduling strategy. The learning rate at epoch \( t \) is computed as:
\[
\text{lr}(t) = \text{lr}_{\min} + \frac{(\text{lr}_{\max} - \text{lr}_{\min})}{2} \left(1 + \cos\left(\frac{\pi t}{T_0}\right)\right),
\]
where \( \text{lr}_{\max} = 1\text{e-}4 \), \( \text{lr}_{\min} = 1\text{e-}6 \), and \( T_0 = 10 \) epochs.

We use the AdamW optimizer with parameters \( \beta_1 = 0.9 \), \( \beta_2 = 0.999 \), and \( \text{weight decay} = 1\text{e-}4 \). To prevent overfitting, we apply a dropout rate of 0.1 in the Transformer layers, label smoothing of 0.1 for classification tasks, and gradient clipping with \( \text{max\_norm} = 1.0 \) to mitigate exploding gradients.

\paragraph{Batching Strategy.}
Considering GPU memory constraints, we set the batch size to 8 during training and 16 during evaluation. To simulate an effective batch size of 32, we use \textbf{Gradient Accumulation} with 4 accumulation steps.

\paragraph{Sequence Configuration.}
The input sequence length is set to 18 time steps, and the prediction length to 24 time steps. With a 1-minute interval between steps, this corresponds to 30 minutes of vessel trajectory data.

\paragraph{Multi-task Loss.}
We define the total loss as a weighted sum of multiple task-specific losses:
\[
\mathcal{L}_{\text{total}} = \lambda_1 \mathcal{L}_{\text{traj}} + \lambda_2 \mathcal{L}_{\text{anom}} + \lambda_3 \mathcal{L}_{\text{coll}} + \lambda_4 \mathcal{L}_{\text{exp}},
\]
where the weights are set as follows: \( \lambda_1 = 2.0 \) (trajectory prediction), \( \lambda_2 = 1.5 \) (anomaly detection), \( \lambda_3 = 1.5 \) (collision risk), and \( \lambda_4 = 1.0 \) (explanation generation).

\paragraph{Early Stopping and Termination.}
We apply early stopping if the validation loss does not improve for 30 consecutive epochs. Training continues for a maximum of 100 epochs but is terminated early if either (i) the validation accuracy reaches a target threshold of 90\%, or (ii) the learning rate decays to the minimum value of \( 1\text{e-}6 \). This configuration ensures stable training while accounting for the complexity and irregularity of maritime data.

\section{C. Training Methodology}
\subsection{C.1 Multi-Task Learning Strategy}
This framework applies a Multi-Task Learning (MTL) strategy to jointly train four tasks: trajectory prediction, anomaly detection, collision risk assessment, and natural language explanation generation. Each task is equipped with an independent prediction head (\texttt{TrajectoryPredictor}, \texttt{AnomalyDetector}, \texttt{CollisionRiskAssessor}), while a shared time-series encoder and a Cross-Modal Alignment module enable inter-task synergy and information sharing.

For trajectory prediction, the model uses the Mean Absolute Error (MAE) loss to enhance robustness against outliers in position errors. In anomaly detection, a weighted Cross Entropy Loss is employed, considering the class imbalance with a ratio of normal:anomaly = 0.3:0.7, along with label smoothing set to 0.1 to improve generalization. Collision risk assessment adopts a Huber Loss, enhanced with a dynamic weighting mechanism that emphasizes high-risk samples. The weight is defined as 
\[
\text{weight} = 1.0 + (\text{target\_risk} \times 2.0),
\]
ensuring that more critical situations receive greater attention during training. For natural language explanation generation, the model utilizes a standard language modeling loss, computed as the average Cross Entropy Loss across all token positions in the sequence. This loss is calculated using the teacher forcing method on concatenated input prompts and target explanations to optimize next-token prediction.

The task-specific weights are set as follows: 
\[
\begin{cases}
\lambda_1 = 2.0 & \text{trajectory prediction} \\
\lambda_2 = 1.5 & \text{anomaly detection} \\
\lambda_3 = 1.5 & \text{collision risk} \\
\lambda_4 = 1.0 & \text{explanation generation}
\end{cases}
\]
with trajectory prediction given the highest importance due to its relevance in maritime safety.

Shared spatio-temporal representations are learned through a Multi-Scale Temporal Attention mechanism with scales of [1, 4, 16, 32], combined with a Cross-Modal Alignment module. An Inverted Embedding approach is used to learn independent temporal patterns for each AIS variable (latitude, longitude, speed over ground, and course over ground), which are shared across all tasks to improve learning efficiency and parameter utilization.

To ensure stable training, gradient accumulation and Xavier uniform initialization are applied. Parameters of the large language model (LLM) remain frozen using pretrained weights. Overfitting is mitigated using early stopping with a patience of 15 epochs, as well as learning rate scheduling.

\subsection{C.2 Self-Supervised Learning Components}
In this framework, a self-supervised learning approach is applied to enable the generation of natural language explanations using a large language model (LLM), even in scenarios where no human-written explanation data is available.

\textbf{Automatic text label generation.} Explanation texts are automatically generated from AIS data by converting numerical features, such as position, speed, heading, anomaly status, and collision risk, into natural language using rule-based methods. This involves denormalizing the values to real-world coordinates and composing comprehensive descriptions that include current and predicted positions, speed changes, course adjustments, and safety analysis. For example, an automatically generated explanation may be: \textit{``Vessel shows normal navigation pattern proceeding from current position (37.5665°N, 126.9780°E) at 15.2 knots toward predicted position (37.5745°N, 126.9820°E) with course adjustment from 45.0° to 48.0°, indicating port approach procedure with appropriate speed reduction to 12.1 knots. Safety assessment reveals low anomaly probability and minimal collision risk, supporting standard maritime operations.''}

\textbf{Teacher forcing-based training.} The LLM decoder is trained using teacher forcing, where the auto-generated explanation serves as the target sequence. Both the prompt and the explanation are generated from the same AIS data input, forming a self-supervised learning setup. The decoder receives the entire input as ``prompt + explanation'' and learns next-token prediction across the full sequence.

\textbf{Prompt-based conditional generation.} During training, a structured prompt beginning with ``\texttt{MARITIME TRAFFIC ANALYSIS}'' is prepended to each instance, followed by the corresponding explanation. This allows the model to learn conditional generation based on context. During inference, dynamic prompts are constructed based on task outputs—trajectory prediction, anomaly detection, and collision risk—enabling the generation of situationally appropriate explanations.

\textbf{Rule-based knowledge injection.} Domain-specific maritime knowledge and the physical semantics of AIS variables are incorporated to ensure high-quality explanations. Patterns in speed, trajectory, and risk level are analyzed to include informative expressions such as ``vessel is accelerating,'' ``normal navigation pattern,'' or ``high collision risk detected.''

\textbf{Cross-modal consistency learning.} The Cross-Modal Alignment module aligns time-series features with natural language outputs by learning attention-based correspondences between time-series embeddings and LLM prompt embeddings. This enhances the semantic consistency between numerical input patterns and their linguistic descriptions, ensuring that generated explanations reflect real-world maritime conditions.

\subsection{C.3 Loss Function Design and Implementation}
\subsubsection{C.3.1 Multi-Task Learning Loss Overview}
This framework adopts a multi-task learning structure that simultaneously trains four tasks: Trajectory Prediction, Anomaly Detection, Vessel Collision Risk Assessment, and Natural Language Explanation Generation. Since each task has different characteristics and output types, task-specific loss functions are designed and combined into a weighted sum.

The total loss function is defined as:
\[
\mathcal{L}_{\text{total}} = \lambda_1 \mathcal{L}_{\text{trajectory}} + \lambda_2 \mathcal{L}_{\text{anomaly}} + \lambda_3 \mathcal{L}_{\text{collision}} + \lambda_4 \mathcal{L}_{\text{explanation}}
\]
The task weights are set to reflect their importance in maritime applications: $\lambda_1 = 2.0$, $\lambda_2 = 1.5$, $\lambda_3 = 1.5$, $\lambda_4 = 1.0$.

\subsubsection{C.3.2 Trajectory Prediction Loss}
The trajectory prediction task is formulated as a regression problem to estimate future vessel coordinates. Due to sudden course changes in maritime environments caused by weather or emergency maneuvers, the Mean Absolute Error (MAE) is employed for its robustness to outliers:
\[
\mathcal{L}_{\text{trajectory}} = \text{MAE}(\hat{y}, y) = \frac{1}{N} \sum_{i=1}^{N} \left| \hat{y}_i - y_i \right|
\]
MAE reduces the influence of extreme values and improves model robustness under unpredictable conditions at sea.

\subsubsection{C.3.3 Anomaly Detection Loss}
Anomaly detection is treated as a binary classification task to distinguish normal and abnormal navigation. In real-world AIS data, normal patterns vastly outnumber anomalies, resulting in a class imbalance problem. To address this, a weighted Cross Entropy Loss with label smoothing is applied:
\[
\mathcal{L}_{\text{anomaly}} = -\sum_{i=1}^{N} w_{y_i} \log(\hat{p}_{y_i})
\]
Class weights are set as $w_0 = 0.3$ (normal) and $w_1 = 0.7$ (anomaly), assigning higher penalty to the minority class. Label smoothing with $\varepsilon = 0.1$ is incorporated to prevent overconfidence on hard labels.

\subsubsection{C.3.4 Collision Risk Assessment Loss}
Collision risk is modeled as a regression task predicting a continuous value between 0 and 1. Because high-risk predictions are critical for maritime safety, the loss function incorporates both robustness and adaptiveness using a weighted Huber Loss:
\[
\mathcal{L}_{\text{collision}} = \frac{1}{N} \sum_{i=1}^{N} w_i \cdot \text{Huber}_\delta(\hat{r}_i, r_i)
\]
The Huber function is defined as:
\[
\text{Huber}_\delta(x) = 
\begin{cases}
0.5 x^2 & \text{if } |x| \leq \delta \\
\delta (|x| - 0.5\delta) & \text{otherwise}
\end{cases}
\]
Adaptive weights are computed as:
\[
w_i = 1.0 + 2.0 \cdot r_i
\]
This design assigns greater penalties to samples with higher risk values $r_i$, encouraging accurate identification of dangerous situations. The threshold $\delta$ is set to 0.1 to ensure L2 behavior for small errors and L1 behavior for large deviations.

\subsubsection{C.3.5 Natural Language Explanation Loss}
Unlike other tasks, explanation generation involves sequential token prediction and is modeled as a language modeling problem. To overcome the lack of manual labels, a self-supervised learning approach is adopted, where auto-generated explanations serve as targets. The standard language modeling loss is used:
\[
\mathcal{L}_{\text{explanation}} = -\sum_{t=1}^{T} \log P(x_t \mid x_{<t}, c)
\]
Here, $x_t$ is the token at time step $t$, and $c$ represents the contextual input derived from other task outputs. The LLM decoder is trained using teacher forcing, where both the prompt and the target explanation are combined into a single input sequence, used as \texttt{input\_ids} and \texttt{labels}. The loss is computed as Cross Entropy over the vocabulary dimension from the LLM's output logits, guiding the model to generate coherent and accurate natural language explanations.

\section{D. Task-Specific Analysis}
\subsection{D.1 Trajectory Prediction}
The objective of trajectory prediction is to forecast the future path of a vessel based on its historical AIS data, which is essential for maritime traffic management and collision avoidance. The input to the model consists of a sequence of 18 AIS time-series points, denoted as $X \in \mathbb{R}^{18 \times 4}$, where each point contains four navigational variables: latitude, longitude, speed over ground (SOG), and course over ground (COG).

Given these 18 input points, the model predicts the vessel’s trajectory for the subsequent 24 points, represented as $\hat{Y} \in \mathbb{R}^{24 \times 4}$. This formulation allows the model to flexibly adapt to various temporal resolutions, depending on the preprocessing strategy applied to the AIS data.

To evaluate prediction performance, we employ metrics tailored to the maritime domain. Average Displacement Error (ADE) measures the mean spatial deviation between the predicted and actual trajectories across all future points, reported in nautical miles. Final Displacement Error (FDE) quantifies the positional error at the final predicted point. We also use Mean Squared Error (MSE) and Mean Absolute Error (MAE) to assess accuracy in predicting vessel speed and heading.

Trajectory prediction inherently involves dynamic modeling that reflects physical influences such as vessel inertia, ocean currents, and wind. In practice, this task supports applications such as route planning, traffic flow forecasting, and port approach coordination.

\subsection{D.2 Anomaly Detection}
The goal of anomaly detection is to identify vessel trajectories that deviate from expected navigational behavior, such as unexpected turns, abrupt speed changes, or route shifts. These anomalies often indicate potential risks or abnormal operations, making their detection critical for maritime safety and surveillance.

To construct a reliable evaluation dataset, we follow the anomaly taxonomy proposed in ~\citet{li2024stad}, which categorizes trajectory anomalies into three types: shift deviation, abnormal heading, and abnormal speed. Based on this taxonomy, we inject synthetic anomalies into normal AIS trajectories by perturbing the relevant navigational variables—latitude, longitude, SOG, or COG—within controlled spatial and temporal bounds. This approach enables the generation of labeled datasets with ground-truth anomaly annotations, facilitating quantitative evaluation.

Given a sequence of AIS points $X \in \mathbb{R}^{T \times 4}$, the model determines whether each sub-segment exhibits anomalous behavior. Model performance is assessed using standard classification metrics, namely Precision and Recall. These metrics reflect both the sensitivity and specificity of the detection approach.

This task supports downstream applications such as vessel behavior monitoring, illegal activity detection, and decision support for maritime authorities. By explicitly modeling navigational patterns and evaluating against labeled anomaly cases, we provide a rigorous foundation for evaluating anomaly detection models in maritime contexts.

\subsection{D.3 Collision Risk Assessment}
The objective of collision risk assessment is to estimate the likelihood of potential collisions between vessels based on their relative motion and spatial proximity. This task plays a pivotal role in ensuring navigational safety, particularly in congested waterways or regions with limited maneuverability.

Following the previous work by~\citet{tritsarolis2023collision}, we define the Collision Risk Index (CRI) as a weighted combination of kinematic factors derived from vessel pairs in encountering processes, including Distance and Time to Closest Point of Approach (DCPA, TCPA), relative bearing, and speed ratio. Using these definitions, we construct a supervised learning dataset by simulating vessel encounters from AIS records and computing the corresponding ground-truth CRI values using established formulas.

Given input features $X \in \mathbb{R}^{N \times d}$ describing the dynamics of vessel pairs—such as their relative distance, speed, course, and encounter geometry—the model learns to predict a scalar CRI value $\hat{y} \in [0, 1]$. The training and evaluation process is carried out using stratified cross-validation over a large-scale real-world AIS dataset curated from the Norwegian coastal region.

Model performance is quantitatively assessed using Mean Absolute Error (MAE) and Root Mean Squared Error (RMSE) against ground-truth CRI values. These metrics reflect both the average deviation and the sensitivity to large errors in risk estimation. Our experiments show that the proposed model achieves significantly lower MAE and RMSE compared to existing methods, highlighting its effectiveness in modeling collision risk from AIS-derived kinematic features.

This task enables practical applications such as real-time risk monitoring, decision support for collision avoidance systems, and predictive safety analysis in autonomous vessel navigation.

\subsection{D.4 Natural Language Explanation Generation}

The objective of natural language explanation generation is to translate complex maritime situation analysis into clear and professional language that is easily interpretable by maritime traffic controllers and vessel operators. This task enhances the transparency of AI-driven decision-making processes, thereby improving user trust and facilitating real-world applicability in safety-critical maritime operations.

To achieve this, prompt templates are carefully designed to reflect standard terminology and procedural conventions in maritime traffic management. The base template follows the structure: \textit{``Based on AIS data analysis from [time\_range], vessel [MMSI] shows [pattern\_description]. Trajectory prediction indicates [future\_movement] with [confidence\_level]. Anomaly assessment: [normal/abnormal] behavior detected. Collision risk: [risk\_level] with nearby vessels.''}

Each sub-task employs a specialized sub-template. For instance, trajectory prediction uses the format: \textit{``will proceed to coordinates [lat, lon] at [speed] knots''}; anomaly detection follows: \textit{``unusual [behavior\_type] detected at [timestamp]''}; and collision risk assessment outputs: \textit{``CRI of [value] indicates [risk\_category] with vessel [target\_MMSI]''}. These structured formats ensure consistency, domain relevance, and clarity across generated explanations.

We evaluate explanation quality using three standard natural language generation metrics: BLEU-4 (n-gram precision), ROUGE-L (longest common subsequence recall), and BERTScore (semantic similarity based on contextual embeddings). Experimental results demonstrate that the model-generated texts accurately capture key maritime semantics while maintaining fluency and task-specific phrasing.

This task contributes to explainable maritime AI by bridging the gap between algorithmic outputs and human interpretability, ultimately supporting informed decision-making in operational maritime environments.

\section{E. Ablation Studies}

\begin{table}[t]
\centering
\resizebox{\linewidth}{!}{%
\begin{tabular}{lcc}
\toprule
\textbf{Configuration} & \textbf{MSE} & \textbf{MAE} \\
\midrule
\textbf{Full Model} & \textbf{95.76} & \textbf{1.96} \\
\midrule
\textbf{Multi-Scale Temporal Attention} & & \\
\hspace{3mm}w/o Multi-scale (Single) & 115.88 & 5.97 \\
\hspace{3mm}Short-term only ([1, 4]) & 101.32 & 4.08 \\
\hspace{3mm}Long-term only ([16, 32]) & 99.49 & 3.91 \\
\midrule
\textbf{Cross-Modality Alignment} & & \\
\hspace{3mm}w/o Cross-modal & 124.11 & 6.54 \\
\hspace{3mm}Simple Concatenation & 103.73 & 3.14 \\
\hspace{3mm}Element-wise Addition & 117.29 & 4.28 \\
\bottomrule
\end{tabular}
} 
\caption{Ablation study of architectural components on the Piraeus AIS dataset.}
\label{tab:ablation}
\end{table}

In this section, we systematically analyze the impact of each component in the proposed AIS-LLM framework. All experiments are conducted on the Piraeus AIS dataset using Mean Squared Error (MSE) and Mean Absolute Error (MAE) as evaluation metrics. The results demonstrate the effectiveness of each architectural component and provide evidence supporting the design choices. Table \ref{tab:ablation} presents the results of a comprehensive study of architectural components.

\subsection{E.1 Multi-Scale Temporal Attention Analysis}
The Multi-Scale Temporal Attention module is a key component for analyzing the complex temporal dynamics of maritime traffic across different scales. In this study, we utilize four temporal scales ([1, 4, 16, 32]) to capture both short-term and long-term navigational patterns. 

Removing the multi-scale mechanism and using only a single temporal scale (w/o Multi-scale) leads to a significant performance drop: MSE increases from 95.76 to 115.88, and MAE increases from 1.96 to 5.97. When using only short-term scales (1, 4), the MSE reaches 101.32, whereas using only long-term scales (16, 32) yields a lower MSE of 99.49, highlighting the importance of long-term patterns in maritime traffic. This degradation emphasizes the necessity of multi-scale temporal reasoning to accommodate both short-term collision avoidance and long-term voyage planning.

\subsection{E.2 Cross-Modality Alignment Effectiveness}
\begin{figure*}[t] 
    \centering
    \includegraphics[width=0.9\linewidth]{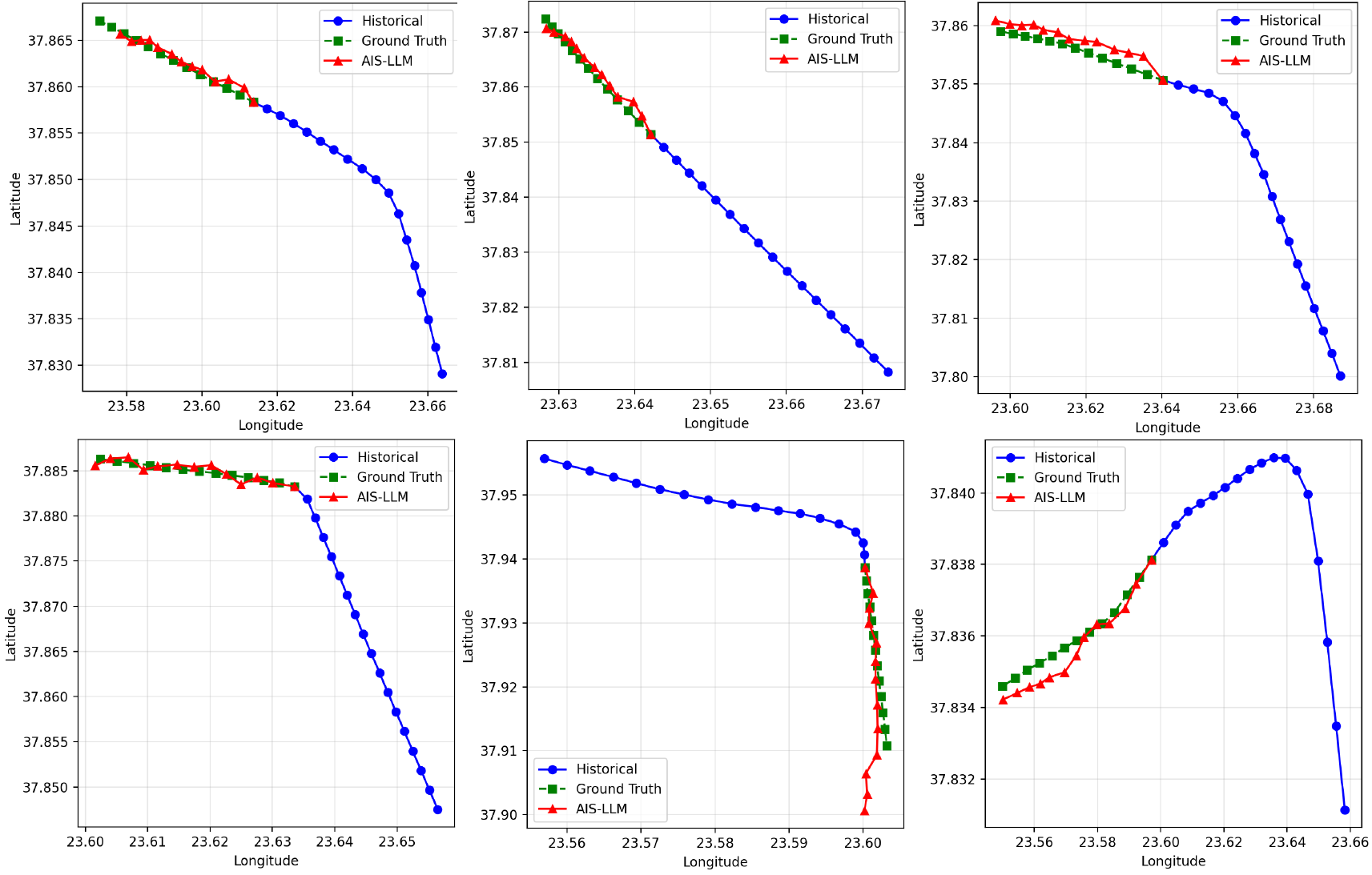}
    \caption{\textbf{Qualitative evaluation results of AIS-LLM Trajectory Prediction for various ship trajectories.}}
    \label{fig:ablation_trajectory_prediction}
\end{figure*}
Cross-Modality Alignment is a core innovation of the framework, enabling effective integration between the temporal patterns of AIS data and semantic knowledge from maritime-domain-specific prompts. To verify the superiority of our attention-based approach, we evaluate different alignment strategies.

When the cross-modal alignment is completely removed (w/o Cross-modal), the model suffers the most severe performance degradation, with MSE increasing to 124.11 and MAE reaching 6.54. This result underscores the critical role of cross-modal alignment and the need to incorporate maritime domain knowledge through textual prompts.

To validate our attention-based design, we compare it with simpler integration approaches. \textbf{Simple Concatenation}, which directly connects time series and text embeddings, achieves MSE 103.73 and MAE 3.14. Although this shows some improvement over no alignment, it fails to capture complex semantic relations between temporal patterns and domain knowledge. \textbf{Element-wise Addition}, which adds the embeddings element-wise, performs even worse (MSE: 117.29, MAE: 4.28), indicating that arithmetic operations cannot model the complex dependencies between heterogeneous modalities.

The superior performance of our attention-based alignment arises from its ability to selectively attend to relevant textual information based on the current temporal context. By using time-series embeddings as queries and prompt embeddings as keys and values, the model dynamically focuses on the most relevant maritime knowledge at each prediction step.

\section{F. Qualitative Results}
\begin{figure*}[t] 
    \centering
    \includegraphics[width=0.85\linewidth]{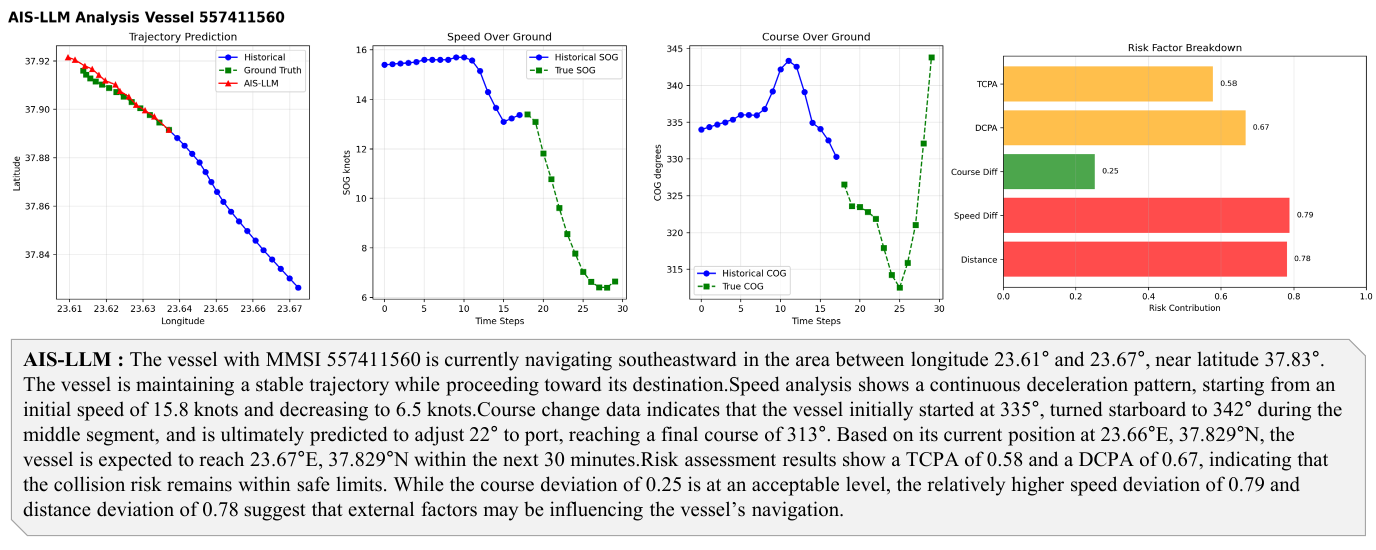}
    \caption{\textbf{Qualitative evaluation results for the description generation function of AIS-LLM (Example 1)}}
    \label{fig:ablation_explain1}
\end{figure*}

\begin{figure*}[t] 
    \centering
    \includegraphics[width=0.85\linewidth]{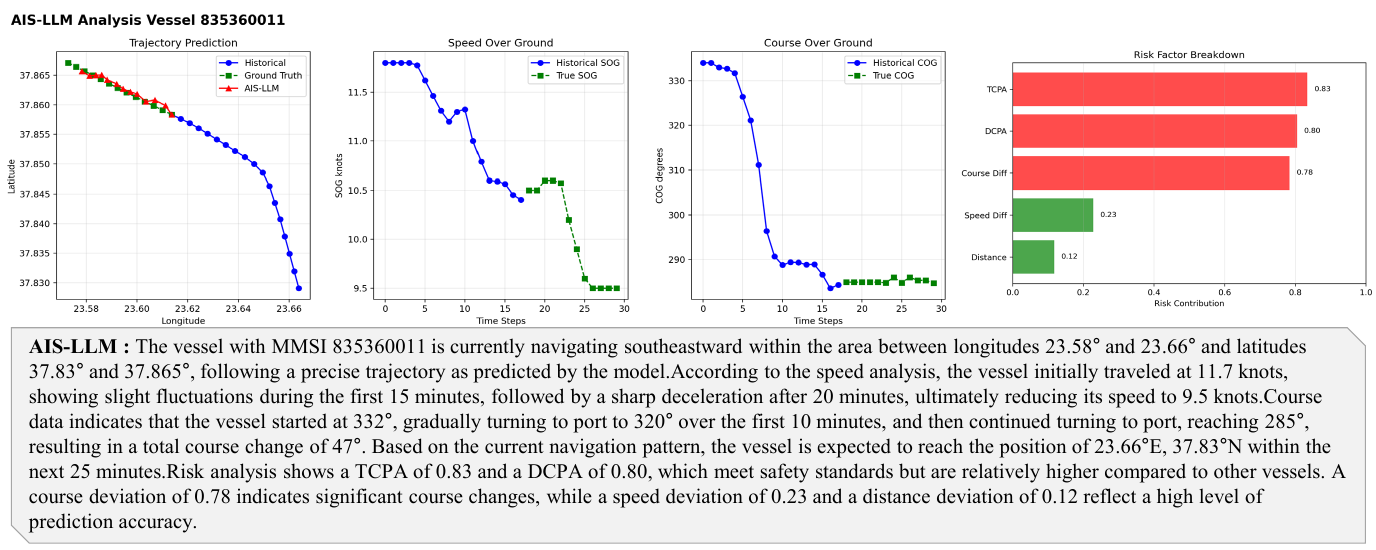}
    \caption{\textbf{Qualitative evaluation results for the description generation function of AIS-LLM (Example 2)}}
    \label{fig:ablation_explain2}
\end{figure*}

\begin{figure*}[t] 
    \centering
    \includegraphics[width=0.85\linewidth]{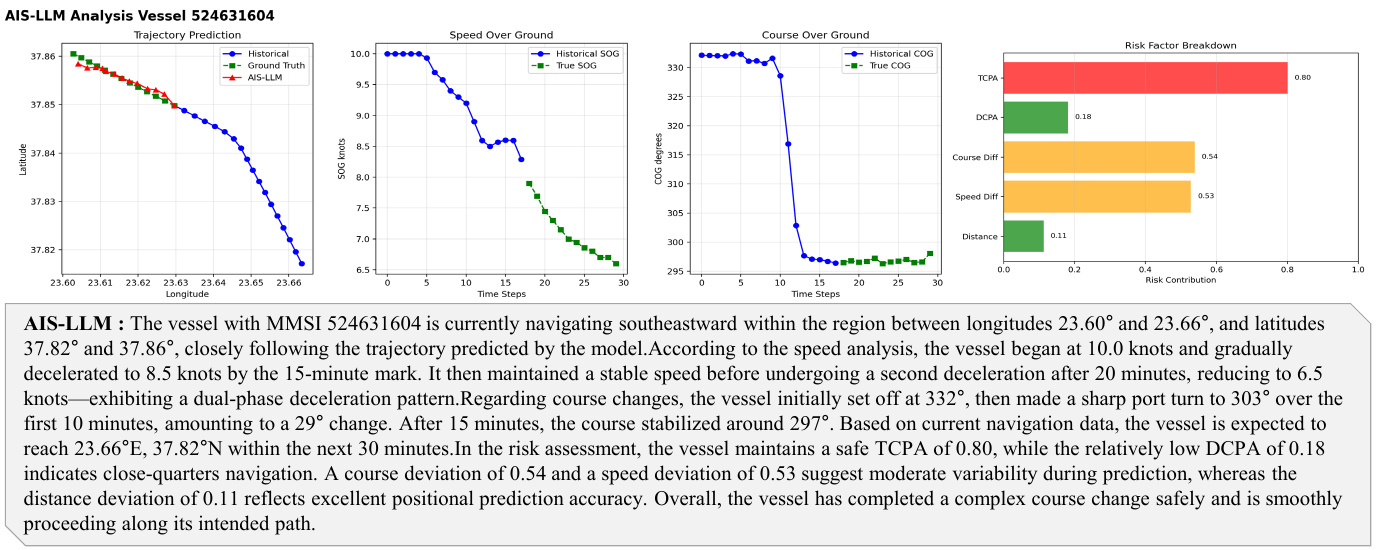}
    \caption{\textbf{Qualitative evaluation results for the description generation function of AIS-LLM (Example 3)}}
    \label{fig:ablation_explain3}
\end{figure*}

This section presents qualitative results to offer a comprehensive understanding of AIS-LLM’s capabilities in both trajectory prediction and natural language explanation. Specifically, Figure~\ref{fig:ablation_trajectory_prediction} demonstrates the model’s performance in forecasting vessel trajectories under various navigational scenarios. Meanwhile, Figures~\ref{fig:ablation_explain1} -~\ref{fig:ablation_explain3} illustrate the model’s ability to generate coherent and contextually appropriate textual explanations.



\end{document}